# Modeling Household Online Shopping Demand in the U.S.: A Machine Learning Approach and Comparative Investigation between 2009 and 2017

Limon Barua[1], Bo Zou[1*], Yan (Joann) Zhou[2], Yulin Liu[3]

[1] Department of Civil, Materials, and Environmental Engineering, University of Illinois at Chicago

[2] Vehicle and Energy Technology & Mobility Analysis, Argonne National Laboratory

[3] Institute of Transportation Studies, University of California, Berkeley

**Abstract:** Despite the rapid growth of online shopping and research interest in the relationship between online and in-store shopping, national-level modeling and investigation of the demand for online shopping with a prediction focus remain limited in the literature. This paper differs from prior work and leverages two recent releases of the U.S. National Household Travel Survey (NHTS) data for 2009 and 2017 to develop machine learning (ML) models, specifically gradient boosting machine (GBM), for predicting household-level online shopping purchases. The NHTS data allow for not only conducting nationwide investigation but also at the level of households, which is more appropriate than at the individual level given the connected consumption and shopping needs of members in a household. We follow a systematic procedure for model development including employing Recursive Feature Elimination algorithm to select input variables (features) in order to reduce the risk of model overfitting and increase model explainability. Among several ML models, GBM is found to yield the best prediction accuracy. Extensive post-modeling investigation is conducted in a comparative manner between 2009 and 2017, including quantifying the importance of each input variable in predicting online shopping demand, and characterizing value-dependent relationships between demand and the input variables. In doing so, two latest advances in machine learning techniques, namely Shapley value-based feature importance and Accumulated Local Effects plots, are adopted to overcome inherent drawbacks of the popular techniques in current ML modeling. The modeling and investigation are performed both at the national level and for three of the largest cities (New York, Los Angeles, and Houston), with a number of findings obtained. The models developed and insights gained can be used for online shopping-related freight demand generation and may also be considered for evaluating the potential impact of relevant policies on online shopping demand.

---

[*] Corresponding author. Email: bzou@uic.edu.

# 1 Introduction

Demand for online shopping is rapidly growing. In the U.S., between 2018 and 2019 the number of online transactions has increased by $76.46 billion, from $523.64 to $600.10 billion. In 2020, U.S. consumers were projected to spend $794.5 billion online, for which part of the growth is due to COVID-19 (Intelligence, 2020). The rapid growth of online shopping has profound impacts on transportation. First, online shopping may substitute, complement, or modify personal travel to stores (Mokhtarian, 2002; Cao, 2009; Shi et al., 2019) and thus have implications for changing personal vehicle miles traveled (VMT). For example, one stream of research argues that reduction in personal VMT as a result of online shopping can be important in low-density areas where travel for shopping takes long distance (e.g., Farag et al., 2003; Goodchild and Wygonik, 2015). An earlier study in the UK estimates that a direction substitution of car trips by delivery van trips could reduce vehicle-km by 70% or more (Cairns, 2005). Yet another stream of research supports a complementary effect, i.e., people frequently buying or searching online tend to make more shopping trips (e.g., Cao, 2012; Zhou and Wang, 2014; Lee et al., 2017). On the other hand, the increase in truck/van traffic for goods deliveries as a result of online shopping growth has raised many concerns, causing greater traffic congestion, shortage in freight parking space, and aggravated road wear-and-tear, particularly in dense urban areas where online shopping demand is high (Crainic et al, 2004; Bates et al., 2018; Jaller and Pahwa, 2020).

To meet the increasing delivery requirements from online shopping – in both volume and delivery speed, logistics service providers have been rethinking and renovating logistics strategies, such as relocating warehouses and expanding the network of distribution centers (Houde et al., 2017; Rodrigue, 2020), employing crowdshipping (Kafle et al., 2017; Hong et al., 2019), and testing drones for contactless delivery (Chiang et al., 2019; Kim et al., 2020). Freight demand derived from online shopping is exerting increasing influences on the planning and operation of freight transportation systems. As such, the ability to predict online shopping demand is important to designing ways to maintain and enhance the performance of freight and overall transportation systems. Moreover, understanding the importance of the input variables used for prediction, and the nature of the dependence of online shopping demand on values of input variables is critical in informing transportation planning and policy-making.

While a body of research has appeared toward understanding online shopping behavior (see Section 2 for a review of the literature), some important gaps remain. First, most of the existing studies focus on the interactive relationship between online and in-store shopping with ample but diverse empirical evidences (Shi et al., 2019). However, the ability to predict the volume of online shopping with reasonable accuracy has not attracted much attention despite its practical importance for transportation planning. Almost all existing research resorts to econometric or statistical models. Based on the reported goodness-of-fit, many of those models would not be adequate for online shopping demand prediction purposes (although prediction is not the main intent of those models). Second, the vast majority of the existing work relies on relatively small and local data samples, lacking a broader understanding of demand pattern from a national perspective. On the other hand, thanks to the recent release of online shopping information in national-level databases, we can build models to learn, on a national scale, how online shopping demand and its influencing factors have evolved over time and vary in different locations. Third, as most of the data in existing work come from surveys of individuals, research related to modeling online shopping demand from the household perspective is insufficient, which should be emphasized as it is more appropriate than at the individual level because online purchases are often made for the needs of the household (e.g., grocery shopping).

This research attempts to fill the above gaps. The contributions of this work are two-folds: empirical and methodological. On the empirical side, this paper leverages two most recent releases of the U.S. National Household Travel Survey (NHTS) – for 2009 and 2017 – to develop machine learning (ML) models for predicting household-level online shopping purchases with input variables encompassing socioeconomic, trip, and land use characteristics and Internet use of household members. We are particularly interested in one type of ML models, gradient boosting machine (GBM), which has several strengths (Friedman, 2001; Elith et al., 2008; Ding et al., 2018, Barua et al., 2020) and shows superior prediction performance in comparison with several alternative prediction techniques for the purpose of the study. After the GBM models are trained, validated, and tested, we further investigate the modeling results by 1) quantifying the importance (i.e., contribution) of each input variable in the models in predicting online shopping demand; and 2) characterizing the relationships between predicted online shopping demand and the input variables. Unlike the existing econometric/statistical approaches which rely on pre-defined model specifications (e.g., linear), our characterization is purely data-driven thus allowing the relationships to vary with input variable values. Results from the investigation are compared between 2009 and 2017, a period in which online shopping has experienced an unprecedent growth, to shed lights on the changes and trends of the factors influencing household online shopping demand. City-specific GBM models are further developed for three of the largest cities in the U.S. (New York, Los Angeles, and Houston), providing additional findings about the commonalities and differences in online purchase characteristics among the three large cities as well as between the cities and the national trends.

The contributions also come from the methodological perspective. First, given a large pool of candidate input variables, a systematic procedure for input variable selection based on Recursive Feature Elimination algorithm is employed to reduce the risk of model overfitting and increase model explainability. Second, in quantifying the importance of each input variables, a recently developed method termed Shapley value-based feature importance (Lundberg and Lee, 2017) is adopted to address possible quantification bias in importance among input variables of different types. Third, instead of using partial dependent plots, a prevalent method for characterizing the relationships between response variable and input variables but can be problematic when input variables are correlated as is often the case, we employ a new approach called Accumulated Local Effects plots developed by Apley and Zhu (2020) that explicitly accounts for the presence of correlation of input variables and is also computationally less expensive.

The remainder of the paper proceeds as follows. Section 2 reviews the relevant literature of online shopping. GBM model development is presented in Sections 3, followed by a description of the data used in the study in Section 4. Section 5 describes model implementation. Section 6 performs post-modeling analysis, including quantifying the importance of the input variables and the relationships between the input and response variables. Section 7 extends the modeling and analysis to three of the largest cities in the U.S. Finally, Section 8 concludes and suggests directions for future research.

## 2 Literature Review

Our review of the literature is organized based on the data used: 1) dedicated survey data for local areas; 2) data as part of a larger travel survey for a metropolitan area; and 3) national-level data. Most of the studies on online shopping behavior are conducted using dedicated surveys conducted at specific locations. Farag et al. (2005) collect a data sample of 826 respondents from four municipalities in the Netherlands to investigate the effects of gender, age, income, land use characteristics, and car ownership on the relationship among frequencies of online searching, online buying, and nondaily shopping trips. Path analysis is conducted. The study is extended by Farag et al. (2007) in which structural equation modeling (SEM) is used. Using data of 392 Internet users from the Columbus metropolitan area in Ohio, Ren and Kwan (2009) estimate a negative binomial and a linear regression model to reexamine the effects of accessibility to local shops and the residential context on the adoption of e-shopping and the frequency of buying online. Age, gender, work hours, income, education, adult percentage in the household, Internet use, race, local population density, and shopping opportunity are included as input variables. Weltevreden and Rietbergen (2007) study the impact of online shopping on in-store shopping based on a dataset of 3,074 Internet users who shop at eight city centers in the Netherlands. The authors use multinomial regression and binomial logistic regression models and find that age, owning a credit card, Internet access and use, and car accessibility value at city centers have significant effects on online shopping. Using data of 539 adult Internet users in the Minneapolis-St. Paul metropolitan area, Cao et al. (2012) investigate the effects of age, the number of vehicles in the household, gender, driving license, income, education, occupation, and employment status on online shopping. It is found that online searching frequency has positive impacts on both online and in-store shopping frequencies and online buying positively affects in-store shopping. For further reviews of the earlier studies, readers may refer to Cao (2009).

Among the more recent research, Lee et al. (2017) use survey data from more than 2,000 residents in Davis, California to explore the effect of personal characteristics, attitudes, perceptions, and the built environment on the frequency of shopping online within three distinct shopping settings. Both univariate ordered response models and pairwise copula-based ordered response models are estimated. The authors find a complementary relationship between online and in-store shopping, even after controlling for demographic variables and attitudes. Using 952 Internet users from two cities in northern California, Zhai et al. (2017) examine the interactions between e-shopping and store-shopping for search goods (books) and experience goods (clothing). The authors find that, among other things, clothing is more likely than books to be associated with store visiting for Internet users. Maat and Konings (2018) investigate whether innovation diffusion or accessibility gains drive the replacement of physical shopping by online shopping, by estimating fractional logit models based on a survey of 534 respondents in Leiden, the Netherlands. Focusing on e-shopping behavior in China, Ding and Lu (2017) use a data sample of 791 respondents from a GPS-based activity travel diary in the Shangdi area of Beijing and develop SEM to investigate the relationships between online shopping, in-store shopping, and other dimensions of activity travel behavior. Similarly, SEM is performed to examine the interaction between e-shopping and in-store shopping using a data sample of 1,032 respondents in the city of Nanjing (Xi et al., 2020). Shi et al. (2019) perform regression analysis using data from interviews with 710 respondents in Chengdu. It is found that e-shopping behavior is significantly affected by sociodemographics, Internet experience, car ownership, and location factors. In addition, the results suggest that e-shopping has a substitution effect on the frequency of shopping trips. The association of spatial attributes with e-shopping is studied in Zhen et al. (2018).

As online shopping is gaining increasing popularity, online shopping information has been incorporated into metropolitan area travel surveys. The use of the information for understanding online shopping behavior is explored by several researchers. Ferrell (2004, 2005) use the San Francisco Bay Area Travel Survey 2000 data to investigate the relationship between home-based teleshopping and shopping travel. In Ferrell (2004), the relationship between travel behavior (number of trips, travel distance, and trip chaining) and home-based teleshopping is explored using linear regression. In Ferrell (2005), the impacts of age, car availability, household income, Internet, homeownership, driving license, education, and health condition of an individual on home-based teleshopping are explored by using SEM. Dias et al. (2020) use the 2017 Puget Sound Household Travel Survey data to explore the relationship between online and in-person engagement in the shopping domain while distinguishing between shopping for non-grocery goods, grocery goods, and ready-to-eat meals. The effects of the number of adults, employment status, population density, household tenure, household type, vehicle availability, and household income on household-level online shopping are explored.

As mentioned in Section 1, due to the scarcity of data and perhaps also unawareness among researchers of the online shopping-related information that has been added to national data sources, national-level research of online shopping behavior remains more limited than studies using dedicated local surveys or metropolitan area travel surveys reviewed above. We are aware of four studies in which national-level datasets are used. Three of them relate to the NHTS data. Zhou and Wang (2014) explore the relationship between online shopping and shopping trips by analyzing the travel pattern-related variables (number of shopping trips, total number of trips, average travel time, gas price) from the 2009 NHTS data. Using the same dataset, Wang and Zhou (2015) develop a binary choice model and a censored negative binomial model to investigate the effects of the Internet, education, age, gender, race, household size, number of household vehicles, home type, population density, rural, and urban size on home delivery frequency. Ramirez (2019) performs negative binomial regression using the 2017 NHTS data to explore the impacts of gender, age, household income, race, education, job category, urban/rural, and the number of drivers in the household on online shopping demand. Besides NHTS data, another national-level data source is the 2016 American Time Use Survey, which is used in Jaller and Pahwa (2020) to investigate the environmental impacts of online shopping. Factors including gender, age, education, employment status, household income, population density, and season are considered to understand their effect on online shopping decisions.

Table 1 summarizes the above reviewed studies with a U.S. focus, given that our interest in this paper is also in U.S. online shopping. In the table, we present the data sources, sample types, and modeling techniques. As is clear in the table and from the review above, all these studies resort to econometric or statistical modeling. Many of the existing studies focus on the relationship between online shopping and in-store shopping, whereas the ability to predict online shopping demand with reasonable accuracy has not been paid attention to despite its importance for transportation planning. Also, econometric/statistical modeling techniques often give an estimate of the effect of an input variable as a single number. However, the effect could vary by the value of the input variable. The constrained, single number-based effect estimates in turn limit the ability of the models to serve demand prediction purposes. Moreover, as online shopping is continuously developing, there is a need but no research for understanding the evolving influence of different input variables on online shopping over time at the national as well as local levels. By leveraging ML and some of its latest advances, our research tries to fill these gaps.

**Table 1** Summary of reviewed U.S.-based online shopping studies

| Studies | Data source | Sample type | Modeling technique |
|---|---|---|---|
| Ren and Kwan (2009) | Survey data from the Columbus metropolitan area | Individual | Negative binomial and linear regression |
| Cao et al. (2012) | Survey data from the Minneapolis-St. Paul metropolitan area | Individual | SEM |
| Lee et al., (2017) | Survey data from Davis, California | Individual | Copula model |
| Zhai et al., (2017) | Survey data from northern California | Individual | Binary logit |
| Ferrell (2004) | 2000 San Francisco Bay Area Travel Survey | Household | Least square regression |
| Ferrell (2005) | 2000 San Francisco Bay Area Travel Survey | Individual | SEM |
| Dias et al. (2020) | 2017 Puget Sound Household Travel Survey | Household | Ordered Probit model |
| Zhou and Wang (2014) | 2009 NHTS | Individual | SEM |
| Wang and Zhou (2015) | 2009 NHTS | Individual | Binary Choice |
| Ramirez (2019) | 2017 NHTS | Individual | Negative binomial regression |
| Jaller and Pahwa (2020) | 2016 American Time Use Survey | Individual | Multinomial Logit |

# 3   Model development

In this study, we develop GBM models for predicting household-level online shopping demand. GBM is a supervised ML technique that repeatedly fits a weak classifier – typically a decision tree, and ensembles the trees to make the final prediction (Regue and Recker, 2014). GBM has several strengths over other ML techniques (Friedman, 2001; Elith et al., 2008; Ding et al., 2018, Barua et al., 2020). First, GBM works very well with high-dimensional mixed-type inputs of numerical and categorical variables. Second, the performance of GBM is invariant to transformations of the input variables and insensitive to outliers. Third, the selection of input variables is internalized in the decision tree, making the algorithm robust to irrelevant input variables. With these strengths, GBM has been reported to yield better prediction than traditional statistical models (e.g., linear regression and ARIMA) and other ML models (e.g., Random Forest (RF), and SVM) on a number of prediction tasks (Ogutu et al., 2011; Zhang and Haghani, 2015).

This section provides a description of the methodology for GBM model development, consisting of three steps: model training, validation, and testing. In line with the three steps, the data used for model development are split into three portions. Following the rule-of-the-thumb (Bisong, 2019), a 60-20-20 split of the data is adopted.

**Step 1:** **Model training.** Use the first portion (60%) of data to train GBM models under different combinations of model hyperparameters.

**Step 2:** **Model validation.** Use the second portion (20%) of data for model validation. This step involves selecting a trained model with the best prediction accuracy but not subject to overfitting.

**Step 3:** **Model testing.** Use the remaining portion (20%) of data to further test the prediction accuracy of the selected GBM model.

## 3.1 Model training

### 3.1.1 Function estimation

Let us use $\{y_i, \boldsymbol{x}_i\}_1^N$ to denote the training sample of known $(y, \boldsymbol{x})$-values, where $y_i$ refers to the response variable and $\boldsymbol{x}_i = (x_i^1, x_i^2, \dots, x_i^d)$ the input variables of the $i$th observation. The goal of model training is to reconstruct the unknown functional dependence $\boldsymbol{x} \xrightarrow{F} y$ with our estimate $\hat{F}(\boldsymbol{x})$, such that the expected value of some specified loss function $L(y, F(\boldsymbol{x}))$ over the joint distribution of all $(y, \boldsymbol{x})$-values is minimized:

$$F^* = \underset{F}{\operatorname{argmin}}\, E_{y,\boldsymbol{x}} L(y, F(\boldsymbol{x})) \tag{1}$$

where $L(y, F)$ is the loss function associated with $y$ and $F$ (e.g., squared error $(y - F)^2$). Thus, the goal of model training can be approximately viewed as minimizing the model prediction error.

The response variable $y$ may come from different distributions. In ML theory, the different distributions naturally lead to different specifications for the loss function $L(y, F)$. Given that online shopping demand is a continuous response variable, the $L_2$ square loss function: $L(y, F)_{L_2} = \frac{1}{2}(y - F)^2$ and the robust regression Huber loss function $L(y, F)_{\text{Huber},\delta}$ are often used (Natekin and Knoll, 2013). We choose the Huber loss function, which captures not only $L_2$ square loss but also mean absolute error $L_1$. As shown in Eq. (2), $L(y, F)_{\text{Huber},\delta}$ is $L(y, F)_{L_2}$ when the absolute error of prediction $|y - F|$ is smaller than or equal to $\delta$, but becomes $L(y, F)_{L_1} = |y - F|$ with a multiplier $\delta$ minus a constant term $\frac{\delta^2}{2}$ when the absolute error of prediction is greater.

$$L(y, F)_{\text{Huber},\delta} = \begin{cases} \dfrac{1}{2}(y - F)^2 & \text{if } |y - F| \leq \delta \\ |y - F|\delta - \dfrac{\delta^2}{2} & \text{if } |y - F| > \delta \end{cases} \tag{2}$$

Following the common procedure in GBM, we parameterize $F(\boldsymbol{x})$ as $F(\boldsymbol{x};\boldsymbol{P})$ where $\boldsymbol{P} = \{P_1, P_2, \dots\}$ is a finite set of parameters. Choosing a parameterized function $F(\boldsymbol{x};\boldsymbol{P})$ then changes to the following problem of parameter optimization:

$$\boldsymbol{P}^* = \underset{\boldsymbol{P}}{\text{argmin}}\, E_{y,x} L\big(y, F(\boldsymbol{x};\boldsymbol{P})\big) \tag{3}$$

Consequently, $F^*(\boldsymbol{x}) = F(\boldsymbol{x};\boldsymbol{P}^*)$.

To determine $\boldsymbol{P}^*$, we employ steepest descent as the numerical minimization method, which iteratively updates $\boldsymbol{P}^*$ as in Eq. (4):

$$\boldsymbol{P}_m = \boldsymbol{P}_{m-1} - \gamma_m \left\{ \left[ \frac{\partial E_{y,x} L(y, F(x;\boldsymbol{P}))}{\partial P_j} \right]_{\boldsymbol{P}=\boldsymbol{P}_{m-1}} \right\} \tag{4}$$

where $P_j$ is the $j$th element in $\boldsymbol{P}$. $\gamma_m$ is obtained from line search as follows:

$$\gamma_m = \underset{\gamma}{\text{argmin}}\, E_{y,x} L\left( y, F\left( \boldsymbol{x};\boldsymbol{P}_{m-1} - \gamma \left\{ \left[ \frac{\partial E_{y,x} L(y, F(x;\boldsymbol{P}))}{\partial P_j} \right]_{\boldsymbol{P}=\boldsymbol{P}_{m-1}} \right\} \right) \right) \tag{5}$$

Note that the minimization problem of (5) only involves one decision variable $\gamma$.

### 3.1.2    Numerical optimization with training data

GBM views each point in $\boldsymbol{x}$ as a "parameter" (so there are $N$ "parameters"). Then, the iterative relationship in steepest descent that corresponds to (4) becomes:

$$F_m(\boldsymbol{x}) = F_{m-1}(\boldsymbol{x}) - \rho_m \left\{ \left[ \frac{\partial L(y, F(x_i))}{\partial F(x_i)} \right]_{F(x)=F_{m-1}(x)} \right\} \tag{6}$$

where $\rho_m = \underset{\rho}{\text{argmin}}\, L\left( y, F_{m-1}(\boldsymbol{x}) - \rho \left\{ \left[ \frac{\partial L(y, F(x_i))}{\partial F(x_i)} \right]_{F(x)=F_{m-1}(x)} \right\} \right)$.

However, there is a key difference here that prevents direct application of the above steepest descent. That is, the gradient is defined only at the data points $\{\boldsymbol{x}_i\}_1^N$ but cannot be generalized to other $\boldsymbol{x}$-values. One way of generalization, according to Friedman (2001), is to parameterize $F(\boldsymbol{x})$ as:

$$F(\boldsymbol{x};\{\rho_m, \boldsymbol{a}_m\}_1^M) = \sum_{m=1}^{M} \rho_m h(\boldsymbol{x};\boldsymbol{a}_m) \tag{7}$$

where $\{\rho_m, \boldsymbol{a}_m\}_1^M$ are parameters. $M$ is the maximum number of iterations in performing the GBM-equivalent steepest descent. The generic functions $h(\boldsymbol{x}; \boldsymbol{a}_m)$, $m = 1,2,\dots,M$ are usually simple parameterized functions of the input variables $\boldsymbol{x}$, characterized by parameters $\boldsymbol{a}_m = \{a_m^1, a_m^2, \dots\}$. In GBM, $h(\boldsymbol{x}; \boldsymbol{a}_m)$ is called a "base learner" and is often a classification tree. In this paper, we consider the following regress trees specification for $h(\boldsymbol{x}; \boldsymbol{a}_m)$:

$$h(\boldsymbol{x}; \boldsymbol{a}_m) = h\left(\boldsymbol{x}; \{b_m^j, R_m^j\}_1^J\right) = \sum_{j=1}^J b_m^j \mathbf{1}(\boldsymbol{x} \in R_m^j) \tag{8}$$

where $\boldsymbol{a}_m = \{b_m^j, R_m^j\}_1^J$. $\{R_m^j\}_1^J$ are disjoint regions that collectively cover the space of all joint values of $\boldsymbol{x}$. These regions are represented by the terminal nodes of the corresponding tree. The indicator function $\mathbf{1}(\cdot)$ takes value 1 if the argument is true, and 0 otherwise. $b_m^j$'s are parameters of the base learner.

Comparing the iterative expression (6) and Eq. (7), the question in the $m$th iteration is to identify $\boldsymbol{a}_m$ such that $h(\boldsymbol{x}; \boldsymbol{a}_m)$ is most parallel to (i.e., most highly correlated with) $\left\{-\left[\frac{\partial L(y, F(x_i))}{\partial F(x_i)}\right]_{F(x)=F_{m-1}(x)}\right\}_1^N$. This can be obtained from the following least-square minimization problem, the reason being that solutions to least-square minimization problems have been well studied and thus can follow standard procedures.

$$\boldsymbol{a}_m = \underset{\boldsymbol{a}, \beta}{\operatorname{argmin}} \sum_{i=1}^N \left[-\left[\frac{\partial L(y, F(x_i))}{\partial F(x_i)}\right]_{F(x)=F_{m-1}(x)} - \beta h(x_i; \boldsymbol{a})\right]^2 \tag{9}$$

The obtained $h(\boldsymbol{x}; \boldsymbol{a}_m)$ is then used to replace $-\left[\frac{\partial L(y, F(x_i))}{\partial F(x_i)}\right]_{F(x)=F_{m-1}(x)}$ in the steepest descent procedure. Specifically, the new line search can be expressed as:

$$\rho_m = \underset{\rho}{\operatorname{argmin}} L(y, F_{m-1}(\boldsymbol{x}) + \nu \rho h(\boldsymbol{x}; \boldsymbol{a}_m)) \tag{10}$$

which is used to update $F(\boldsymbol{x})$:

$$F_m(\boldsymbol{x}) = F_{m-1}(\boldsymbol{x}) + \nu \rho_m h(\boldsymbol{x}; \boldsymbol{a}_m) \tag{11}$$

where $\nu \in (0,1]$ is the learning rate, a hyperparameter in the GBM model. Considering a learning rate less than one attempts to prevent overfitting by "shrinking" the update of $F(\boldsymbol{x})$. Previous numerical experiments revealed that a small $\nu$ can result in better prediction performance of GBM models.

In ML, the process represented by (9)-(11) is called "boosting". The overall procedure thus gets the name of "gradient boosting". Overall, the GBM algorithm can be summarized as follows:

| GBM Algorithm |
|---|
| 1. **Initialization:** $F_0(\boldsymbol{x}) = \mathrm{argmin}_\rho \sum_{i=1}^N L(y_i, \rho)$; set hyperparameter values |
| 2. **For** $m = 1$ to $M$ do: |
| 3. $\quad \boldsymbol{a}_m = \underset{\boldsymbol{a}, \beta}{\mathrm{argmin}} \sum_{i=1}^N \left[ -\left[ \frac{\partial L(y, F(x_i))}{\partial F(x_i)} \right]_{F(x)=F_{m-1}(x)} - \beta h(\boldsymbol{x}_i; \boldsymbol{a}) \right]^2$ |
| 4. $\quad \rho_m = \underset{\rho}{\mathrm{argmin}} \, L(y, F_{m-1}(\boldsymbol{x}) + \nu \rho h(\boldsymbol{x}; \boldsymbol{a}_m))$ |
| 5. $\quad F_m(\boldsymbol{x}) = F_{m-1}(\boldsymbol{x}) + \nu \rho_m h(\boldsymbol{x}; \boldsymbol{a}_m)$ |
| 6. **End for** |

Note that four hyperparameters are involved in GBM model training: the number of regression trees ($M$), the maximum depth of a tree, the minimum sample leaf of a tree (i.e., the minimum number of observations a node needs to have to be considered for splitting), and the learning rate ($\nu$). Grid search is performed to enumerate possible value combinations for the four hyperparameters. Each combination results in one trained GBM model.

## 3.2 Model validation

Model validation consists of identifying the combination of hyperparameter values that yields the best model fit without overfitting. To select the GBM model with the highest prediction accuracy, $R^2$ is used:

$$R^2 = 1 - \frac{\sum_{i=1}^N (y_i - \hat{y}_i)^2}{\sum_{i=1}^N (y_i - \bar{y})^2} \tag{12}$$

where $y_i$ denotes the observed value of the $i^{\text{th}}$ observation, $\hat{y}_i$ is the corresponding predicted value, $\bar{y}$ is the mean of the observed values: $\bar{y} = \frac{1}{n} \sum_{i=1}^N y_i$.

We calculate $R^2$ for each trained model and sort the models in descending order based on $R^2$. These models are then evaluated one by one starting from the one with the highest $R^2$, as follows. We apply a trained model to the validation dataset to generate predicted values and calculate $R^2$. If the difference between this $R^2$ and the $R^2$ associated with the training dataset is less than a threshold (0.1 in this study), then the model is selected as the best model. Otherwise, the difference in $R^2$ suggests presence of overfitting. Then the model is discarded and the next model for evaluation is studied. In the end, the best combination of hyperparameter values, which correspond to the first encountered model without overfitting, is identified.

To further assure that the selected hyperparameter values lead to a good GBM model, $k$-fold cross validation is also performed. Specifically, the training and validation datasets are merged and randomly divided into $k$ subsets. Then, $k - 1$ subsets are selected for training a GBM model using the selected hyperparameter values. The trained model is then used for prediction using the remaining subset. $R^2$'s of the training subset and the testing subset are calculated. This process is repeated $k$ times. If the average $R^2$ associated with model validation is much lower than with model training, then the hyperparameter values are discarded. The next best combination of hyperparameter values (based on description of the previous paragraph) is evaluated. Otherwise, the selected hyperparameters and associated GBM model are kept.

### 3.3 Model testing

Given the selected GBM model, the model testing step is to provide an understanding about how accurate the model prediction could be on new data. Specifically, after model training and validation, the remaining 20% of the data not used in the previous two steps are used to check if the model can still yield good accuracy in prediction. In measuring the prediction accuracy, root-mean-square error (RMSE) is used in addition to $R^2$. As shown in Eq. (13), RMSE is defined as the square root of the average of squared differences between predicted and observed values over all observations. A lower RMSE value means a smaller average difference between $y_i$ and $\hat{y}_i$, thus a better fit of the model.

$$RMSE = \sqrt{\frac{1}{N'}\sum_{i}^{N'}(\hat{y}_i - y_i)^2} \tag{13}$$

where $\hat{y}_i$ and $y_i$ are the predicted and observed values of the $i$th observation. $N'$ is the testing data size.

## 4 Data

The NHTS data of 2009 and 2017 are used in this study (FHWA, 2012; 2017). NHTS data are collected from a stratified random sample of U.S. households providing detailed information on individual- and household-level travel behavior along with socioeconomic, demographic, and geographic factors that influence travel decisions. The 2009 NHTS survey interviewed 150,147 households which include 308,901 individuals. The 2017 NHTS survey interviewed 129,696, which include 264,234 individuals. A new feature of 2009 and 2017 NHTS data that does not exist in previous versions is the inclusion of information on the number of online purchases made by an individual that are delivered to home in the month prior to the survey date. Since the focus of this study is at the household level, we aggregate online purchases of individuals in the month prior to the survey date to the household level, and use household online purchases as the response variable. If the online purchase record for a member in a household is missing, then the household is not included in our dataset. After aggregation, the distributions of household online purchases in 2009 and 2017 are shown in Fig. 1.

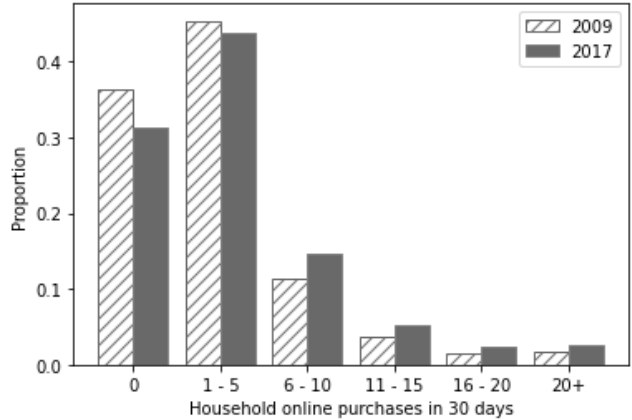

**Fig. 1.** Distribution of household online purchases in 2009 and 2017 (source: NHTS data)

Besides the new information on individual online purchases, the 2009 and 2017 NHTS data also contain rich information about individual- and household-level socioeconomic, travel, and other characteristics. Specifically, the NHTS data consist of four data files for both 2009 and 2017: household file, person file, vehicle file, and trip file. As their names imply, each file contains a different set of variables. In this study, these files are merged and processed to generate household-specific variables. The final dataset used for this study contains four categories of household-level variables: socioeconomic characteristics, trip characteristics, land use characteristics, and Internet use.

As the starting point, we consider 48 candidate input variables for each year. The full lists of the variables are provided in Appendices 1 and 2. Note that some minor differences exist in the list of variables between 2009 and 2017, due to the differences in data provision from NHTS. In 2009, four categorical variables about house type (duplex; townhouse; apartment or condominium; and mobile home or trailer) are included, but not the 2017 list. On the other hand, unique in the 2017 are: 1) percentage of members in a household with excellent health conditions; 2) percentage of members in a household with poor health conditions; 3) indicator of whether all household members use smartphones daily; and 4) indicator of whether all household members use laptop/desktop daily. The extent to which these variables affect online purchases will be examined along with other candidate variables that are common in 2009 and 2017 NHTS data in subsection 5.1.

# 5 Model implementation and results

## 5.1 Input variable selection

Given the large number (48) of candidate input variables, it will be desirable to build less complex models with fewer features, by deciding which input variables are essential for prediction and which are not. This can be useful when one wants to reduce the risk of overfitting and increase model explainability (Guyon et al., 2002; Burkov, 2019). To this end, some feature selection procedure needs to be performed. The idea is to discard input variables that make limited contributions to model predictability. In this paper, we consider Recursive Feature Elimination (RFE) algorithm, which requires moderate computation efforts (Guyon et al., 2002) and is shown to perform better than other feature selection techniques such as least absolute shrinkage and selection operator (LASSO) and principal component analysis (PCA), especially in the case where input features demonstrate strong nonlinear, interactive, or polynomial relationships (Xue et al., 2018).

RFE recursively removes one feature at a time with the least importance, retrains the model, re-ranks the remaining features, and then removes the next feature with the least importance. Initially, for each of the two years we perform model training and validation as described in subsections 3.1 and 3.2 (without performing $k$-fold cross validation) to come up with the best GBM model with the full list of 48 candidate input variables. A 10-fold cross validation is then performed on the model. The average $R^2$ from applying the model to 10 different testing subsets is recorded, termed as the 10-fold cross validation score of the model. Then, the importance of each feature is computed following the procedure described later in subsection 6.1, based on which we remove the feature with the lowest feature importance. We start the next iteration and repeat the same procedure, with 47 input variables. The iterations continue until the 10-fold cross validation scores of models from two consecutive iterations is greater than a predefined threshold (stopping criteria), which we set as 0.01. Summarizing, RFE algorithm is represented as follows:

---

RFE Algorithm

---
1. **Initialization**: Data $(y, \boldsymbol{x})$
2. **Repeat**
3.     **Train and identify** the best trained GBM model using $(y, \boldsymbol{x})$
4.     **Compute** 10-fold cross validation score for the model
5.     **Determine** feature importance
6.     **Identify and remove** input variable $x'$ with the least importance
7.     **Update** input variables $\boldsymbol{x} \leftarrow \boldsymbol{x} - x'$
8. **Until** stopping criteria is met

---

After implementing RFE algorithm, 16 input variables are retained for both 2009 and 2017. Interestingly, the 16 input variables are the same for both years, as listed in Table 2 below. Table 3 provides summary statistics of these variables.

**Table 2:** Variable categories, names, and definitions

| Category | Variable name | Definition |
|---|---|---|
| Response variable | Online purchases | Number of purchases over the Internet by a household in the last 30 days from the survey date |
| Socioeconomic characteristics | Average member age | Average age of household members |
| | Male percentage | Percentage of male members in a household |
| | Household size | Number of household members |
| | Household income | Household annual income (in $000) |
| | Adult percentage | Percentage of adults (age $\geq$ 18) in a household |
| | No high school percentage | Percentage of household members without a high school degree |
| | Bachelor's degree percentage | Percentage of household members with a bachelor's degree |
| | Number of vehicles | Number of vehicles in a household |
| | Home ownership | Indicator of whether a household owns the home property |
| Trip characteristics | Number of trips per day | Number of trips made by all household members in a travel day |
| | Travel time per day | Total travel time of all household members in a travel day |
| | Gas price | Gas price in a travel day |
| | Shopping trip percentage | Percentage of shopping trips in total trips made by a household in a travel day |
| Land use characteristics | Urban area | Indicator of whether a household lives in an urban area |
| | Population density | Population density in the census tract of the household location |
| Internet use | Daily Internet use | Indicator of whether all household members use the Internet daily |

Note: In NHTS data, household income and population density are recorded as ranges. We take the middle of the corresponding range as the value for each observation.

**Table 3:** Descriptive statistics of variables

| Category | Variable | Type | 2009 | | | | 2017 | | | |
|---|---|---|---|---|---|---|---|---|---|---|
| | | | Min | Max | Mean | Std. dev. | Min | Max | Mean | Std. Dev. |
| Response variable | Online purchases | Continuous | 0 | 265 | 3.4 | 6.3 | 0 | 198 | 4.2 | 6.4 |
| Socioeconomic characteristics | Average member age | Continuous | 18 | 92 | 52.7 | 14.2 | 18 | 92 | 57.6 | 23.8 |
| | Male percentage | Continuous | 0 | 100 | 44.2 | 33.2 | 0 | 100 | 45.2 | 34.6 |
| | Household size | Continuous | 1 | 14 | 2.6 | 1.3 | 1 | 10 | 1.7 | 0.7 |
| | Household income | Discrete | 5 | 100 | 64.4 | 29.7 | 10 | 2,000 | 72.7 | 53.2 |
| | Adult percentage | Continuous | 11.1 | 100 | 86.2 | 21.7 | 33.3 | 100 | 89.1 | 5.4 |
| | No high school percentage | Continuous | 0 | 100 | 2.7 | 14.3 | 0 | 100 | 3.7 | 15.9 |
| | Bachelor's degree percentage | Continuous | 0 | 100 | 25.2 | 37.4 | 0 | 100 | 24.8 | 37.7 |
| | Number of vehicles | Binary | 0 | 1 | 0.9 | 0.3 | 0 | 1 | 0.8 | 0.4 |
| Trip characteristics | Number of trips per day | Continuous | 1 | 52 | 7.4 | 4.7 | 1 | 60 | 7.3 | 4.5 |
| | Travel time per day | Continuous | 0 | 27 | 2.3 | 1.1 | 0 | 12 | 1.9 | 1.2 |
| | Gas price | Continuous | 0.2 | 1230 | 21.8 | 30 | 0.4 | 2,040 | 42.9 | 60.1 |
| | Shopping trip percentage | Continuous | 149.5 | 446 | 285.6 | 94.5 | 201.3 | 295.1 | 240.3 | 22.8 |
| | Number of trips per day | Continuous | 0 | 100 | 23.8 | 25.5 | 0 | 100 | 9.9 | 13.9 |
| Land use characteristics | Urban area | Binary | 0 | 1 | 0.1 | 0.3 | 0 | 1 | 0.1 | 0.3 |
| | Population density | Discrete | 50 | 30,000 | 3,146.4 | 4,577 | 50 | 30,000 | 3,770 | 5,365 |
| Internet use | Daily Internet use | Binary | 0 | 1 | 0.8 | 0.4 | 0 | 1 | 0.9 | 0.3 |

## 5.2   Prediction performance of the GBM models

With the 16 input variables, two GBM models are developed, one for 2009 and the other for 2017. Table 4 shows $R^2$ values from applying the GBM models to training, validation, and testing datasets. Also reported in the table are the mean and standard deviation of $R^2$ values associated with the testing subsets in cross validation (in parentheses). The generally high $R^2$ values indicate good fit of the trained models. We observe small differences between the mean $R^2$ values from cross validation and the $R^2$ values using the training datasets. The $R^2$ values using the testing datasets are also high, suggesting a high level of prediction accuracy when the models are applied to new datasets.

**Table 4:** $R^2$ of the GBM models when applying to different datasets

| Datasets | Training | Validation | Cross validation | Testing |
|---|---|---|---|---|
| 2009 | 0.77 | 0.73 | 0.73 (0.05) | 0.72 |
| 2017 | 0.75 | 0.70 | 0.70 (0.04) | 0.71 |

The RMSE values presented in Table 5 tell a similar story. Smaller RMSE values are obtained for 2009 and 2017 datasets. Considering that the number of online purchases ranges from 0 to 198, an RMSE of 2.78 for 2009 and 2.93 for 2017 using the testing datasets corroborate the good performance of the trained GBM models when applied to new datasets.

**Table 5:** RMSEs of the GBM models when applying to different datasets

| Datasets | Training | Validation | Testing |
|---|---|---|---|
| 2009 | 2.57 | 2.76 | 2.78 |
| 2017 | 2.83 | 2.98 | 2.93 |

To further examine the prediction performance of the GBM models, we compare the models with several alternative models including linear regression, quadratic regression, SVM, and RF, using the same response and input variables. Specification of the linear regression model is straightforward. For quadratic regression, the input variables along with their squared and cross-product terms are included. In developing SVM and RF, tuning hyperparameters is critical. For SVM, three hyperparameters (kernel, regularization parameter, and kernel coefficient) need to be tuned. Three types of kernel functions are tested: linear, polynomial, and radial basis functions. Regularization parameter is tuned from 0.1 to 1000. Kernel coefficient is tested from 0.0001 to 1. After tuning, we find that the radial basis function kernel with a regularization parameter of 10 and a kernel coefficient of 0.01 yields the highest $R^2$ using the testing data for both 2009 and 2017. For RF, three hyperparameters to be tuned are: the number of trees, the maximum number of features used for splitting a tree, and the minimum sample leaf size (the minimum number of samples required for a leaf node). A large number of combinations of hyperparameter values are tested. We tune the number of trees from 1 to 1000, the maximum number of features from 1 to 16 (the number of input variables), and the minimum sample leaf size from 1 to 40. We find that RF models with 450 trees, a maximum number of features of 16, and a minimum sample leaf size of 25 yield the highest $R^2$ using the testing data for both 2009 and 2017.

Table 6 compares the $R^2$ and RMSE values of linear regression, quadratic regression, SVM, RF, and GBM using the same testing data. The results indicate that the order of performance is: linear regression < quadratic regression < SVM < RF < GBM. The order is consistent for both years and under both $R^2$ and RMSE. The results corroborate the superior prediction performance of GBM.

**Table 6.** $R^2$ and RMSE for different prediction models

| Prediction models | $R^2$ | | RMSE | |
|---|---|---|---|---|
| | 2009 | 2017 | 2009 | 2017 |
| Linear regression | 0.11 | 0.11 | 13.28 | 14.70 |
| Quadratic regression | 0.13 | 0.12 | 11.93 | 12.04 |
| SVM | 0.59 | 0.56 | 4.55 | 5.16 |
| RF | 0.66 | 0.65 | 3.64 | 3.76 |
| GBM | 0.72 | 0.71 | 2.78 | 2.93 |

# 6    Post-modeling analysis

In our study, we aim to gain an understanding of the influence of the input variables and their interactions on online purchases. In this section, we use Shapley values to quantify the importance of each input variable in predicting online purchases of a household and the accumulated local effects plots to interpret the underlying relationships between different input variables and online purchases.

## 6.1    Quantifying importance of input variables

### 6.1.1    Method

To quantify the importance of input variables, the most commonly employed method for tree-based models (such as GBM) is based on Gini importance (Strobl et al., 2007; Zhou and Hooker, 2019), for which the relative importance of an input variable in each decision tree is the sum of improvements in the squared error from the splits involving the input variable (Hastie et al., 2009). The relative importance is then averaged over all decision trees to obtain the relative importance of the input variable. However, the Gini importance has a drawback. It is known to be biased towards input variables with continuous and discrete variable with high cardinality (Zhou and Hooker, 2019; Aldrich, 2020; Gómez-Ramírez et al, 2020), as these variables provide high possibilities for tree splitting. To address this issue, Lundberg and Lee (2017) propose a method that is based on Shapley values (Hur et al., 2017; Aldrich, 2020). Stemming from game theory, Shapley values provide a theoretically justified way to fairly allocate a coalition's output among members in the coalition (Shapley, 1953). In the context of this paper, coalition members are input variables which collectively produce the GBM model output. The Shapley value-based method is adopted to quantify importance of the input variables.

In calculating the Shapley values, it is assumed that coalition members join a game in sequence. The sequence of joining is important especially when members may have similar skills. Conceptually, if two members have overlapping skills, then the member joining the game earlier is expected to make greater contribution to the coalition's output than the other member who joins later. In view of this, Shapley values characterize each member's contribution to the coalition's output as the averaged value over every possible sequence of coalition members. Now we apply this idea to quantifying input variable importance. Consider that a GBM model has $d$ input variables. Let $\boldsymbol{x}^i = (x_1^i, x_2^i, \dots, x_d^i)$ denote the value of the input variables for the $i$th observation. Each input variable in the observation is viewed as a coalition member. The contribution of the input variable $j$ of observation $i$, termed Shapley value $\emptyset_j^i(v)$, is calculated as:

$$\emptyset_j^i = \sum_{S \subseteq \left\{x_1^i, \dots, x_{j-1}^i, x_{j+1}^i, \dots, x_d^i\right\}} \frac{|S|! \, (d - |S| - 1)!}{d!} \left(F\left(S \cup \left\{x_j^i\right\}\right) - F(S)\right) \tag{14}$$

In Eq. (14), $S$ can be any subset of the full set of input variables excluding $x_j^i$. $F(\cdot)$ denotes the trained GBM model. $F\left(S \cup \left\{x_j^i\right\}\right)$ is trained with input variables being $S \cup \left\{x_j^i\right\}$, and $F(S)$ is trained with input variables being $S$. The difference of $F\left(S \cup \left\{x_j^i\right\}\right)$ and $F(S)$ then provides an indication of the contribution of $x_j^i$ to the predicted value of $F(\cdot)$. The contribution is weighted considering the sequence of input variables, for which $x_j^i$ is placed in the $(|S| + 1)$th place and the calculation of contribution only involves input variables up to $x_j^i$ in the sequence. Thus, given subset $S$ which is placed at the beginning of the sequence followed by $x_j^i$, there are $|S|! \, (d - |S| - 1)!$ possible sequences. On the other hand, the total number of all possible sequences is $d!$. Thus, the weight is $\frac{|S|!(d-|S|-1)!}{d!}$. We then sum over all possible subsets to obtain the Shapley value of input variable $j$ of observation $i$.

The Shapley value of input variable $j$, which is denoted as $I_j$ and measures the importance of the variable, is obtained by summing $\emptyset_j^i$'s over all observations:

$$I_j = \sum_{i=1}^{N} \emptyset_j^i(v) \tag{15}$$

### 6.1.2 Results

By applying the Shapley value-based method, the importance of all input variables in the GBM models for 2009 and 2017 is computed with results displayed in Fig. 2 (ranked based on importance in 2017). We also present the change in the ranking of importance of the input variables between 2009 and 2017 in Fig. 3, where blue arrows indicate no change in ranking, red arrows denote ranking drops, and green arrows represent ranking rises. For the discussions below, we focus on the ranking and ranking changes of the input variables.

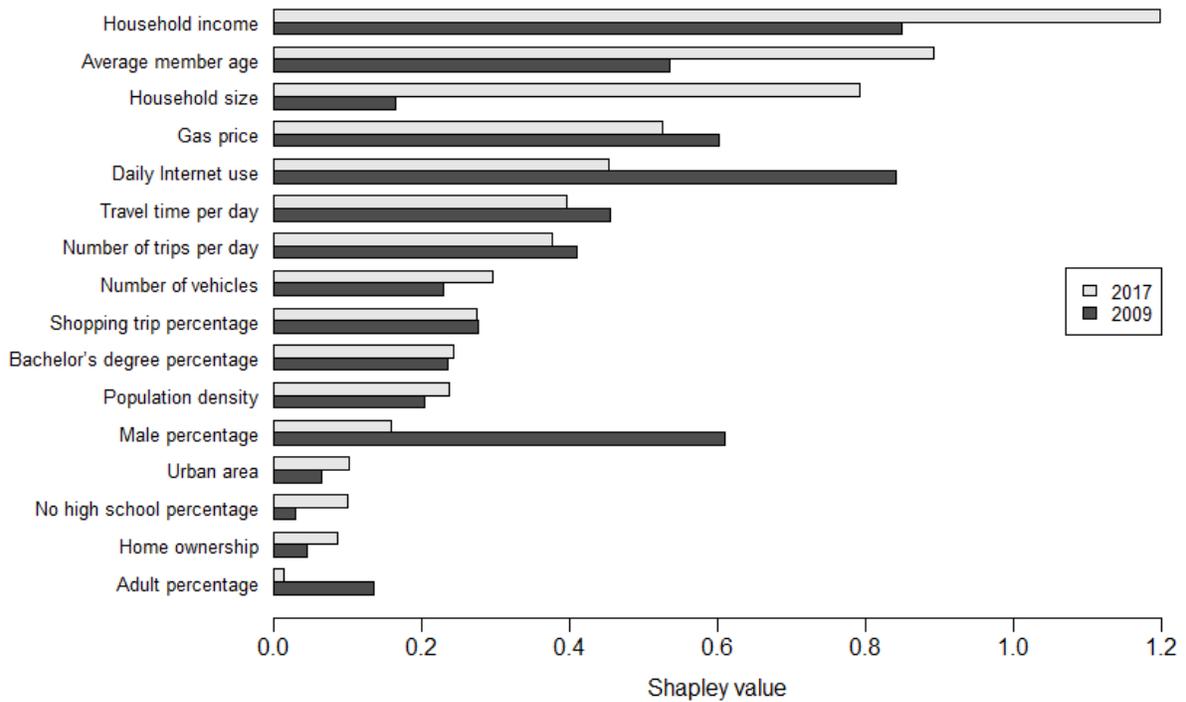

**Fig. 2.** Importance of input variables

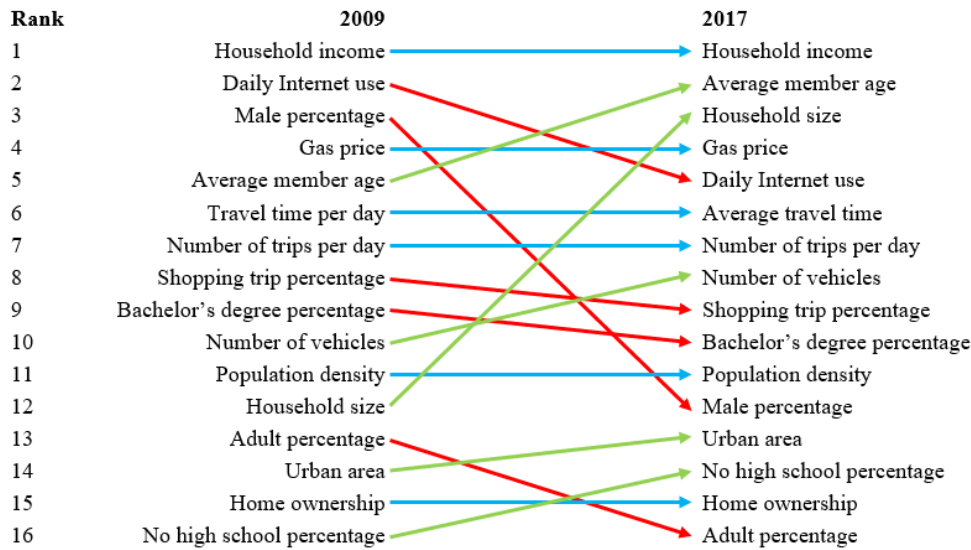

| Rank | 2009 | 2017 |
|------|------|------|
| 1 | Household income | Household income |
| 2 | Daily Internet use | Average member age |
| 3 | Male percentage | Household size |
| 4 | Gas price | Gas price |
| 5 | Average member age | Daily Internet use |
| 6 | Travel time per day | Average travel time |
| 7 | Number of trips per day | Number of trips per day |
| 8 | Shopping trip percentage | Number of vehicles |
| 9 | Bachelor's degree percentage | Shopping trip percentage |
| 10 | Number of vehicles | Bachelor's degree percentage |
| 11 | Population density | Population density |
| 12 | Household size | Male percentage |
| 13 | Adult percentage | Urban area |
| 14 | Urban area | No high school percentage |
| 15 | Home ownership | Home ownership |
| 16 | No high school percentage | Adult percentage |

**Fig. 3.** Change in the ranking of input variable importance from 2009 to 2017

Household income is the most important variable for online shopping for both 2009 and 2017, with a higher Shapley value in 2017. The variable that indicates whether all household members use the Internet daily is the second most important variable in 2009, whereas its importance drops to the fifth place in 2017. This may suggest that people in 2009 depended more on daily Internet use in making online shopping decisions than in 2017. As the Internet has become widespread over time, it is reasonable to see the decline in the importance of the daily Internet use.

Besides household income and daily Internet use, the percentage of male members in a household is the third most important input variable in 2009, whereas its importance is very low in 2017. The difference may be explained from the perspective that technology use related to online shopping was evaluated more differently by gender in 2009, as supported by prior research (Venkatesh and Morris, 2000). On the other hand, with continuous penetration of the Internet in people's lives, its acceptance among women has increased between 2009 and 2017 thus largely filling the gender gap in Internet use (Morahan-Martin, 2009; Pew Research Center, 2019). As a result, we observe much lower importance of gender in 2017. Other than gender, gas price ranks fourth in both years with a similar Shapley value. Gas price is important as it affects the cost of going to stores for shopping. Also, people of different ages may have quite different tendency for shopping online. As such, it is not surprising that the average age of household members is the second most important variable in 2017, though in 2009 it ranks fifth.

The ensuing input variables in the 2009 ranking are mostly related to household trip characteristics, including total travel time of all household members per day, the number of trips made by a household per day, percentage of shopping trips, and the number of vehicles in a household. As both travel and online shopping consume time and online shopping can be competing and/or complementary with traveling to stores for shopping, characteristics related to trip-making are obvious predictors of online purchases. Fig.

3 shows that the four input variables have quite consistent importance between the two years. Travel time per day and the number of trips per day are the sixth and seventh most important features in both 2009 and 2017, with close Shapley values. The percentage of shopping trips ranks eighth in 2009, while its importance goes down to the ninth in 2017. The importance of the number of vehicles in a household has increased between the two years, with a higher Shapley value in 2017.

Besides male percentage, the other most significant change in Shapley value ranking occurs to household size, from the 12th in 2009 to the 3rd in 2017, with the Shapley value increased from 0.18 to around 0.78. This may be explained by the fact that online shopping has become a routine for households in 2017 compared to 2009. As a result, the number of online purchases in a household is critically dependent on the size of the household. Population density holds the eleventh place in both years with similar Shapley values. Adult percentage, urban area, home ownership, and no high school percentage have even smaller importance in predicting household online purchases for both 2009 and 2017.

## 6.2   Understanding the relationships between input and response variables

While Shapley values provide a single number for each input variable in a model to represent the importance of the input variable in driving model prediction, more investigation is needed if one wants to further understand how predicted online purchases are affected by input variables at different values. For this purpose, partial dependence plots (PDP) has been most commonly used, which is a graphical rendering of the predicted response variable value as a function of one or multiple input variables while accounting for the average effects of the other input variables (Friedman, 2001; Zhao and Hastie, 2019). PDP works by marginalizing the model response over the distribution of the variables other than the input variable under evaluation (Hastie et al., 2009). For input variable $j$ of observation $l$ ($x_{l,j}$), its partial dependence value is calculated as $\bar{f}_j(x_{l,j}) = \frac{1}{N}\sum_{i=1}^{N} f(x_{l,j}, x_{i,\setminus j})$, where $N$ is the total number of observations and $x_{i,\setminus j}$ is the vector of values for the other input variables of observation $i$.

An underlying, often untested assumption of PDP is that the variable under evaluation is not correlated with the other input variables, which is a strong assumption and presents a serious issue because input variables almost always bear some degree of correlation, as is our case (Appendix 3 presents the correlation matrices for the 16 input variables for 2009 and 2017). To make this more clear, in the equation above some combinations of $x_{l,j}$ with $x_{i,\setminus j}$ can result in artificial data instances that are unlikely based on the actual observations (e.g., a household has a very low income and a very large number of vehicles), which biases the estimated input variable effect (Molnar, 2019).

To address this issue, this study adopts a recently developed technique, termed Accumulated Local Effects (ALE) plot (Apley and Zhu, 2020), as an alternative. In addition to accounting for correlation among input variables, ALE plots are computationally less expensive than PDP (Apley and Zhu, 2020). The construction of the ALE estimator for an input variable proceeds as follows. Let $x_{i,j}$ denote a continuous input variable $j$ of observation $i$. $\boldsymbol{x}_{i,\backslash j}$ represents the remaining input variables of observation $i$. To calculate ALE of input variable $j$, the value range of $\{x_{i,j}: i = 1,2, \dots, N\}$ (in total $N$ observations) is partitioned into $K$ intervals: $(z_{k-1,j}, z_{k,j}]: k = 1,2, \dots, K$ where $z_{k,j}$ is the $(k/K)$-quantile value of the empirical distribution of $\{x_{i,j}: i = 1,2, \dots, N\}$. $z_{0,j}$ is chosen just below the smallest observed $x_{i,j}$ value, and $z_{K,j}$ chosen the largest observed $x_{i,j}$ value, for input variable $j$. We let $n_j(k)$ denote the number of observations that fall into the $k$th interval $(z_{k-1,j}, z_{k,j}]$. For a particular observation $x_{l,j}, l = 1,2, \dots, N$ for input variable $j$, let $k_j(x_{l,j})$ denote the index of the interval into which $x_{l,j}$ falls, i.e., $x_{l,j} \in (z_{k_j(x_{l,j})-1,j}, z_{k_j(x_{l,j}),j}]$.

With the above notations, we first compute the uncentered ALE $\hat{g}_{j,ALE}(x_{l,j})$ for $x_{l,j}$:

$$\hat{g}_{j,ALE}(x_{l,j}) = \sum_{k=1}^{k_j(x_{l,j})} \frac{1}{n_j(k)} \sum_{\{i: x_{i,j} \in (z_{k-1,j}, z_{k,j}]\}} \{f(z_{k,j}, \boldsymbol{x}_{i,\backslash j}) - f(z_{k-1,j}, \boldsymbol{x}_{i,\backslash j})\} \qquad (16)$$

In Eq. (16), $f(z_{k,j}, \boldsymbol{x}_{i,\backslash j}) - f(z_{k-1,j}, \boldsymbol{x}_{i,\backslash j})$ is the difference of the predicted response variable value for observation $i$, when input variable $j$ takes the upper and lower bounds of interval $k$: $(z_{k-1,j}, z_{k,j}]$. We sum over all observations $i$'s of which $x_{i,j}$ falls into this interval, and divide the sum by the number of observations in the interval $n_j(k)$. Thus, $\frac{1}{n_j(k)} \sum_{\{i: x_{i,j} \in (z_{k-1,j}, z_{k,j}]\}} \{f(z_{k,j}, \boldsymbol{x}_{i,\backslash j}) - f(z_{k-1,j}, \boldsymbol{x}_{i,\backslash j})\}$ gives the averaged incremental effect of input variable $j$ changing from $z_{k-1,j}$ to $z_{k,j}$. We then sum the incremental effects over all intervals up to the one to which $x_{l,j}$ falls into, to obtain the accumulated effect of $x_{l,j}$.

For a continuous input variable, the actual value in ALE plots is demeaned, i.e., the value of Eq. (16) is reduced by the mean value. Eq. (17) gives the centered ALE $\hat{f}_{j,ALE}(x_{l,j})$ for $x_{l,j}$:

$$\hat{f}_{j,ALE}(x_{l,j}) = \hat{g}_{j,ALE}(x_{l,j}) - \frac{1}{N} \sum_{k=1}^{K} n_j(k) \cdot \hat{g}_{j,ALE}(z_{k,j}) \qquad (17)$$

In the above ALE computation, the correlation is accounted for by partitioning the value range of input variable $j$ into $K$ intervals and considering combinations of $z_{k,j}$ and $z_{k-1,j}$ with only observed $\boldsymbol{x}_{i,\backslash j}$ values from the correspoinding interval. This largely avoids unrealistic combinations of $x_{l,j}$ with $\boldsymbol{x}_{i,\backslash j}$ values that are not observed in the data. Note that if the input variable of interest $j$ is a binary variable, then there will be just one interval $\left[z_{0,j}, z_{1,j}\right]$ for Eq. (16), where $z_{0,j} = 0$ and $z_{1,j} = 1$. Demeaning (Eq. (17)) is not needed. Interested readers may refer to Apley and Zhu (2020) for further theoretical details.

In what follows, we present the ALE plots in four subsections (6.2.1-6.2.4) each corresponding to one category of input variables shown in Table 2. In each category, the input variables are arranged in the order of their feature importance in 2017 (shown on the right column in Fig. 3).

### 6.2.1  Socioeconomic characteristics

The ALE plots for input variables in the socioeconomic characteristics category is presented in Fig. 4. For household income, we observe that in 2009 online shopping purchases of a household slightly decreases when household income increases from $2,500 to around $15,000 and then increases more monotonically. For 2017, a more homogeneous increasing trend is observed. The drop in online purchases as household income increases at the beginning may be a reflection of the preference of low-income households for in-store shopping. As income increases, greater affordability for transportation could prompt households to switch from online to in-store shopping, though the effect is quite small. On the other hand, the positive relationship of online purchases with household income is intuitive, consistent with prior empirical evidences (e.g., Ferrell, 2005; Wang and Zhou, 2015; Lee et al., 2017), and can be attributed to three factors. First, more affluent households tend to purchase more (either online or in stores). Second, the time value of more affluent households is higher. Everything else being equal, such households tend to prefer the option of online shopping which demands less of their time. Third, more affluent households are less sensitive to additional cost of shipping than households with lower income.

The average age of household members has an overall negative effect on online purchases in both 2009 and 2017, which reflects the fact that young people are more interested than more senior people in online shopping. This can be explained by the fact that computer use skills, which are essential for online shopping, are more easily learned among younger people, as suggested by earlier work (Czaja et al., 1989; Hernández et al. 2011). In addition, young people usually possess greater experience with the Internet, and their attitude toward using new technology holds greater importance in decision-making processes related to technology adoption (Morris and Venkatesh, 2000). In contrast, earlier research reveal that more senior people perceive the Internet with greater risks and place more importance on the perception of self-efficacy (Trocchia and Janda, 2000) – in this context, shopping without relying on the Internet.

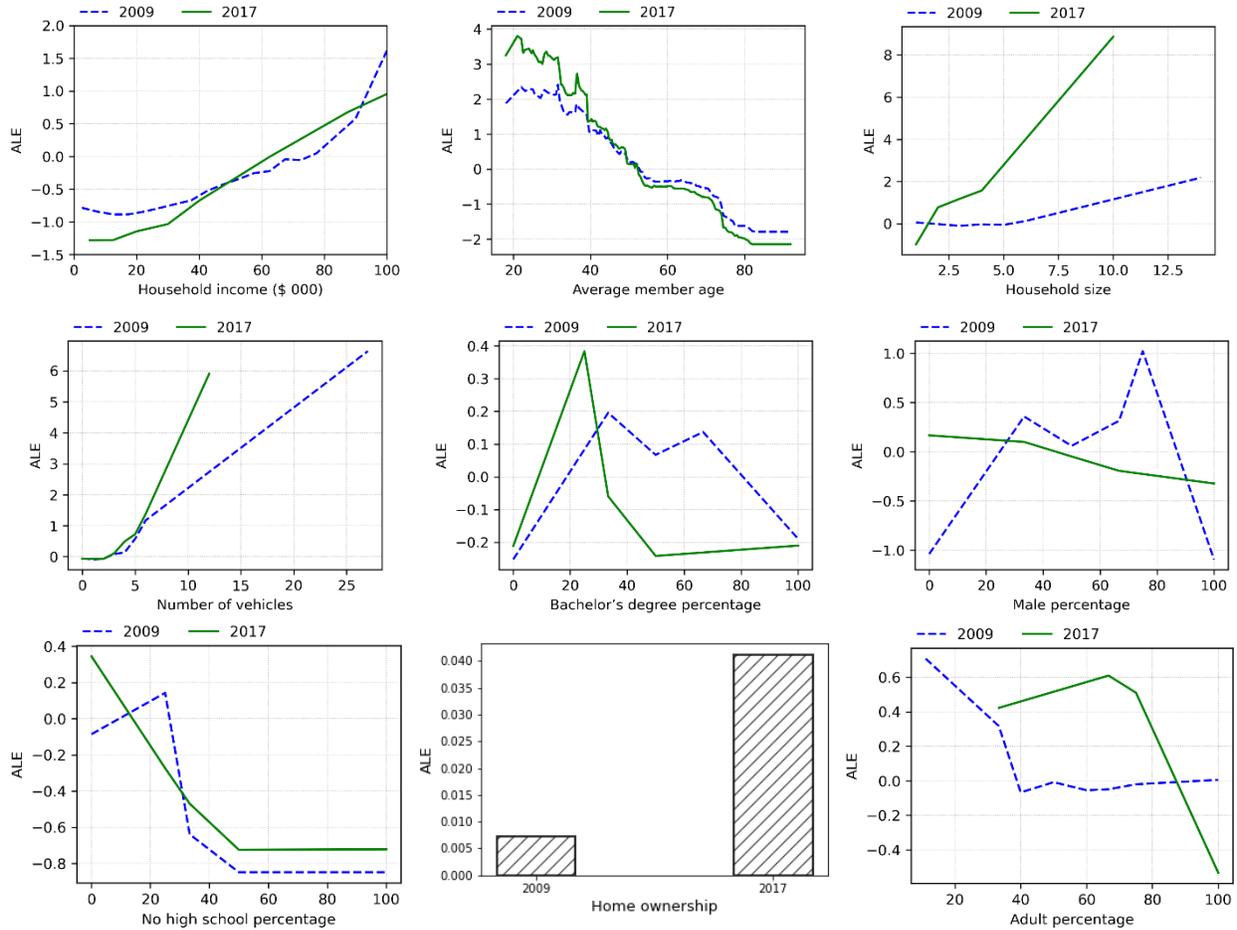

**Fig. 4.** ALE plots for input variables in the socioeconomic characteristics category

The number of online purchases has a positive relationship with household size, except for the household size below five in 2009 for which online purchases are almost invariant to household size. The generally positive relationship is not surprising: more people in a household typically means higher demand for shopping. Everything else being equal, this will translate to more online purchases. For most household sizes, the number of online purchases is greater in 2017 than in 2009, supporting the argument that online shopping has gained greater popularity in 2017. Online purchases tend to be invariant to the number of vehicles in a household – up to four vehicles in 2009 and two vehicles in 2017 – and then increase with the number of vehicles. Since most households in the dataset have no more than four vehicles (the percentage is 96% for 2009 and 97% for 2017), the ALE plot suggests that online purchases are not sensitive to vehicle ownership for most households. For the positive relationship when the number of vehicles is large, a possible explanation is that these households could be engaged in vehicle-dependent or related businesses, e.g., second-hand car sale, auto lease or rental. The vehicles are typically not used for household shopping purposes. In addition, with a large vehicle fleet to handle, a household engaged in vehicle businesses may have a constrained schedule for in-store shopping.

Turning to the two education related variables, the percentage of household members with a bachelor's degree does not give a clear-cut message. In both years, the highest online purchases occur when a household has part of its members with a bachelor's degree. While some prior investigations support that higher education increases one's Internet use capability, which enables and encourages online shopping (Farag et al., 2007; Cao et al., 2012), the non-monotonic relationship found here is more in line with the arguments in other existing research that education background has no, negative, or mixed effects on online shopping and that online shopping is actually a relatively easy task that does not require higher education (Mahmood et al., 2004; Zhou et al., 2007). Nonetheless, we speculate that some basic Internet literacy is still needed. A too shallow education background may still affect a household's propensity for online shopping. This is supported by the overall negative relationship between online purchases and the percentage of household members without a high school degree. It is also interesting to observe that the lowest propensity is achieved when members without a high school degree dominate a household (50%), and remain the same low level as the percentage increases.

The gender (im)balance and adult percentage in a household show some interesting results. In 2009, a household with a more balanced male/female composition tends to have the highest Internet purchases. On the other hand, the role of gender largely diminishes in 2017, with a slight trend that online purchases decrease with greater male percentage in the household, which is consistent with the feature importance results in subsection 6.1.2 and with findings in prior work (Lee et al., 2015; Hernández et al., 2011). For the percentage of adults in a household, in 2009, online purchases decrease with adult percentage, up to 40%. A possible explanation is that at that time non-adults, especially teenagers might be more familiar with online shopping than adults. After eight years, in 2017 those Internet-versed teenagers had grown up as adults, leading to the change in the trend. For the decline of online purchases in the range of 80-100% adults, it may be reflective of a large number of such households consisting of senior/retired household members, who are more traditional and still go to stores for shopping.

Finally, the ALE plot shows that owning the home property tends to encourage online purchases. The difference is even amplified in 2017 compared to 2009. A possible reason for the renting-owning difference is that owning a home property (e.g., owning a single-family house as opposed to renting an apartment unit) gives a household a sense of permanency and possibly more space (a single-family house is likely to be larger than an apartment unit), and consequently makes the household purchase more to improve the living place (buying appliances, decorations, etc.), whereas such motivation would be less if just temporarily renting a place.

### 6.2.2   Trip characteristics

The ALE plots for input variables in the trip characteristics category are presented in Fig. 5. First, the ALE of gas price shows some interesting results. In 2009, online purchases decrease when gas price increases from $1.5/gallon to around $2.25/gallon, and then stay roughly constant when the gas price is between $2.25/gallon and about $4.0/gallon. But online purchases start to increase as gas price goes beyond $4.0/gallon. The initial decline seems counterintuitive at first sight. A possible explanation, following Ma et al. (2011), is that as the initial gas price increases from a low base price, the dominant factor affecting online purchases may be the reduction in the budget allocatable for shopping, which leads to a decline in online purchases. On the other hand, the increasing trend when gas price is over $4.0/gallon is understandable: as gas price increases, driving becomes more expensive, adding to the generalized travel cost to go to stores. Consequently, online shopping becomes more attractive. In contrast to 2009, the overall trend of online shopping varies less in 2017 over a narrower range of gas price, though with some fluctuations. The difference in the range coverage of gas price in the two years is due to less variation of gas price in 2017 than in 2009. In general, households seem to be less sensitive to gas price when purchasing online in 2017.

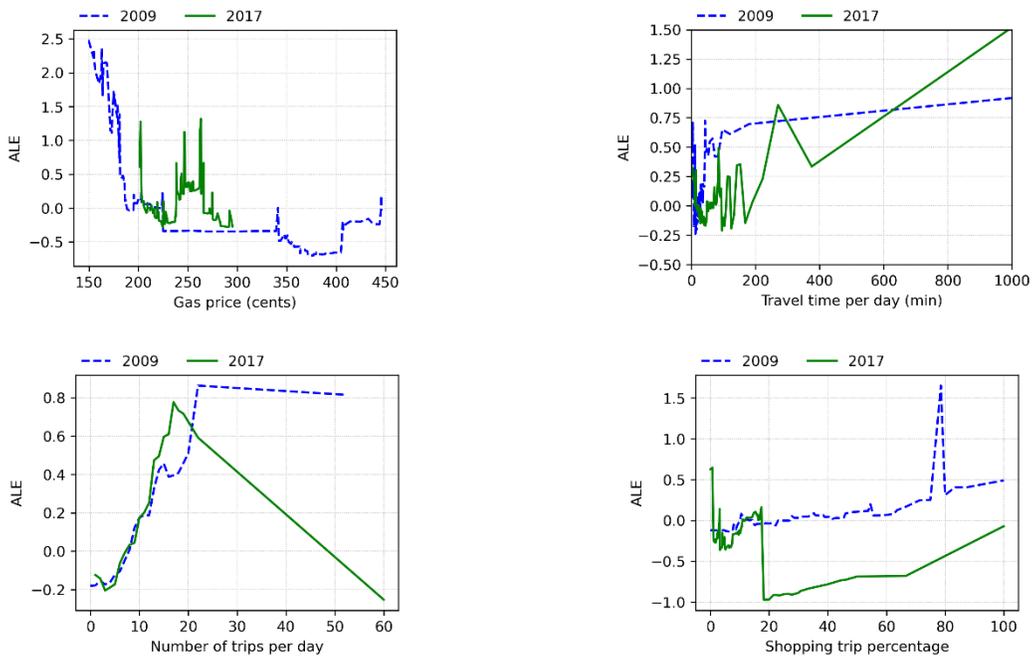

**Fig. 5.** ALE plots for input variables in the trip characteristics category

Turning to the ALE plot for travel time of household members per day, the two curves for 2009 and 2017 both follow an overall increasing trend. As a household spends more time traveling, it is likely to have less time available for shopping. As online shopping requires less time and activities than in-store shopping, household members with less shopping time are naturally more inclined to purchase over the Internet. We also note that when household travel time is near zero, the ALE values are actually not, or even close to the lowest. Our speculation is that people with almost no travel at all will spend most of the time at home, thus likely taking care of things including shopping through the Internet as much as possible.

Following the same argument as for the time use by trips, online purchases are positively related with the number of trips made by a household in a day in 2009. In 2017, the increase continues up to about 17 trips, after which online purchases start to decline. A possible explanation is that more trips could involve buying things from stores on the way (although the main purpose of such trips is not necessarily shopping), thus reducing the need for online shopping. Online purchases with respect to the percentage of shopping trips follows a more consistent increasing trend in 2009 (up to about 22 trips per day), which supports a broad claim of a complementary association between online and in-store shopping as found in earlier work (e.g., Farag et al., 2005; 2007; Cao et al., 2012; Lee et al., 2017; Xi et al., 2020). On the other hand, a sudden drop is observed in 2017 when shopping trip percentage is around 20%, which may suggest the existence of substitution at some point as a household increases shopping percentage in total trips. Also, as shopping trips change from zero to non-zero, some online purchases would likely be substituted by in-store buying. This effect seems more evident for 2017.

### 6.2.3   Land use characteristics

The ALE plots for the input variables in the category of land use characteristics are presented in Fig. 6. For population density, we observe a "V" shape, or a first-decreasing-then-increasing trend, which can be explained as follows. When population density is very low, it probably would require a long trip to get to a nearby store for shopping. In this case, shopping over the Internet would be more convenient saving households a substantial amount of shopping-related travel time. As population density increases, the time spent in going to stores is decreased. As a result, households will be more willing to shop in stores. As population density continues to increase, households again become more inclined to online shopping, which may be attributed to two factors. First, greater population density means greater human interactions in working, social, and other contexts, reducing the time available for in-store shopping. Second, previous research has argued that people living in dense areas tend to have greater access to the Internet (Loomis and Taylor, 2012), which is essential to online shopping. Related to this, households in an urban location tend to shop more than in non-urban areas. Between the two years, the effects of population density and urban location are stronger in 2017 than in 2009.

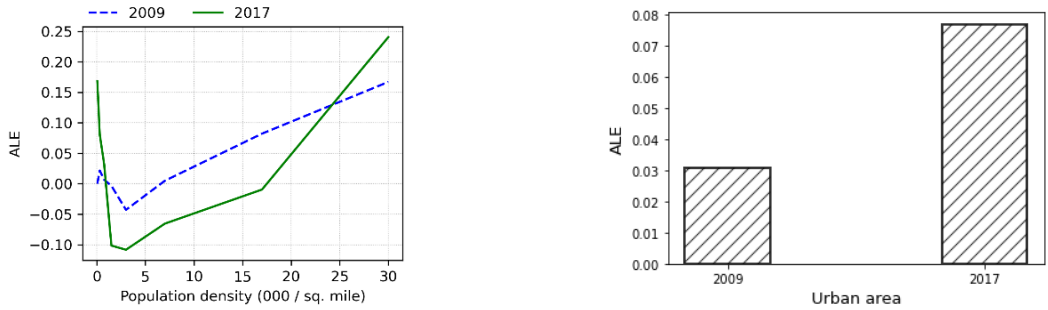

**Fig. 6.** ALE plots for input variables in the land use characteristics category

### 6.2.4 Internet use

The ALE plot for the binary input variable indicating whether all members in a household use the Internet daily is presented in Fig. 7. Since the variable is binary, ALE is presented in two bars for each year, one with daily Internet use and the other without. The plot clearly shows that daily Internet usage has a significant impact on online purchases for both 2009 and 2017, which supports the argument that more frequent use of the Internet enables more online shopping. This may also be attributed to additional online shopping demand that is "induced" from more frequent Internet use, a phenomenon that has been seen in other transportation contexts (e.g., Cervero and Hansen, 2002; Zou and Hansen, 2012). With daily Internet use, the average number of online purchases in a household in a 30-day period will be about 1.1 higher than otherwise. In 2017, the difference is slightly smaller (about 0.9).

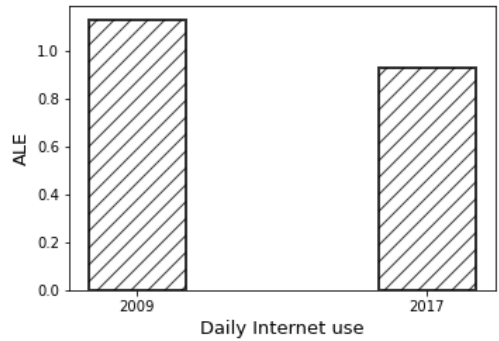

**Fig. 7.** ALE plot for household daily Internet use

## 7   Further analysis on three large cities

In this section, we further investigate ALE plots for three of the four largest cities in the U.S. – New York, Los Angeles, and Houston. We exclude Chicago since there are insufficient records in the NHTS data (the data size for Chicago is only about 10% of that for New York in 2009 and 15% in 2017). We follow the same procedure of model training, validation, and testing described in Section 3, now using city-specific data. Thus, six city-specific GBM models are trained. The data size (number of household observations), best hyperparameter values, and $R^2$ of the GBM models based on the testing data are reported in Table 7. We observe that the $R^2$'s of the city-specific models are lower than the national-level models (in Table 4). This is not surprising and attributed to a much smaller number of data points for training each city-specific model than training the national models.

**Table 7:** Data size, hyperparameter values, and model fit of trained city-specific models

|  | New York | | Los Angeles | | Houston | |
|---|---|---|---|---|---|---|
|  | 2009 | 2017 | 2009 | 2017 | 2009 | 2017 |
| Data size | 2,827 | 2,992 | 2,922 | 1,840 | 2,601 | 2,709 |
| Number of regression trees | 400 | 400 | 450 | 300 | 500 | 450 |
| Max depth of tree | 7 | 8 | 6 | 7 | 8 | 6 |
| Min sample leaf size | 10 | 12 | 12 | 8 | 10 | 10 |
| Learning rate | 0.01 | 0.02 | 0.02 | 0.04 | 0.02 | 0.03 |
| $R^2$ (based on testing data) | 0.52 | 0.51 | 0.54 | 0.51 | 0.56 | 0.54 |

Using the developed city-specific models, ALEs are plotted with respect to each of the input variables, for each city and for 2009 and 2017. We find that overall, the ALE plots follow the trends of the national models in subsection 6.2, with some interesting new findings for some or all of the three big cities. These new findings are presented below (Fig. 8-15). Each figure corresponds to one input variable, with the left graph plotting ALE for 2009 and the right graph plotting ALE for 2017.

For input variables in the socioeconomic characteristics category, the first interesting finding occurs to household income (Fig. 8). Compared to the national ALE, online purchases seem largely invariant to household income from about $7,000 to $40,000 in 2009. Similarly, for household income of Houston between about $6,000 and $30,000 in 2017. For all three cities, online purchases are rather insensitive to household size in 2009 (Fig. 9). In 2017, the insensitivity remains for Los Angeles and Houston for household size between two and five (six). The relationship for New York follows more the national trend. It is interesting to see that for all three cities, a two-member household has significantly more online purchases than a single-member household in 2017, which is different from 2009. Fig. 10 shows that the number of vehicles of a household seems to affect online purchases in these big cities mostly with a limit, which is unlike the national level (Fig. 4) when the number of vehicles in a household becomes large. In 2009, the effect is plateaued in Los Angeles and Houston after the number of vehicles reaches six. In 2017, that number is four for New York (roughly plateaued), seven for Los Angeles, and six for Houston.

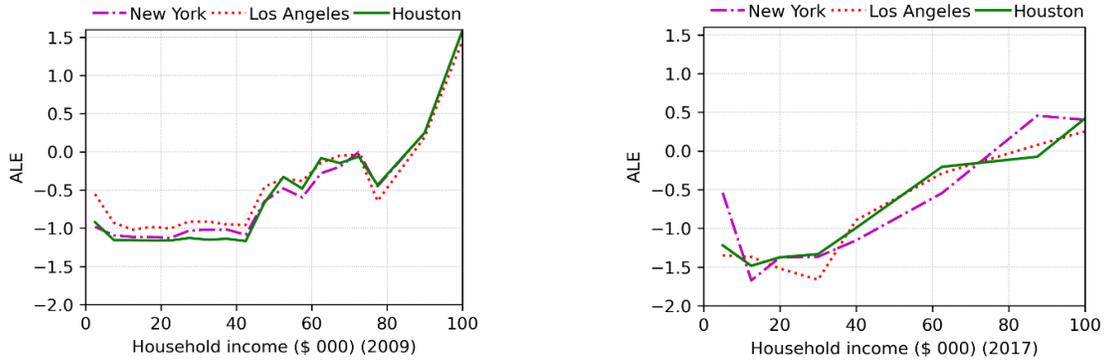

**Fig. 8.** ALE plot for household income of the three cities

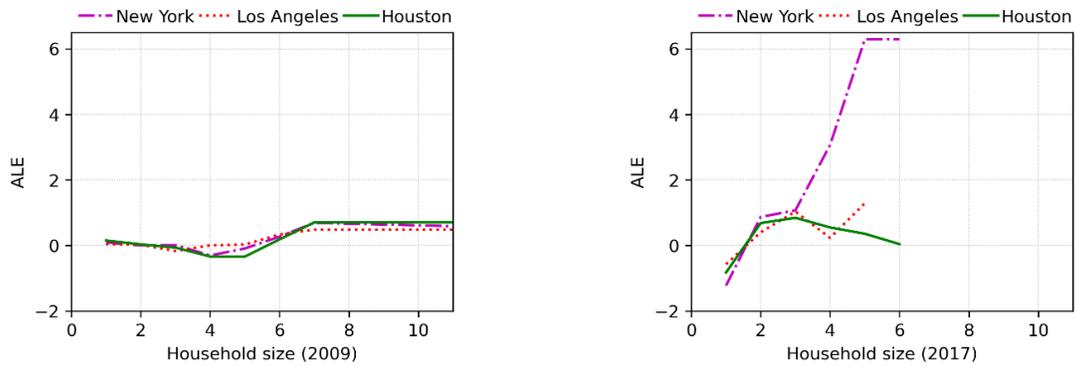

**Fig. 9.** ALE plot for household size of the three cities

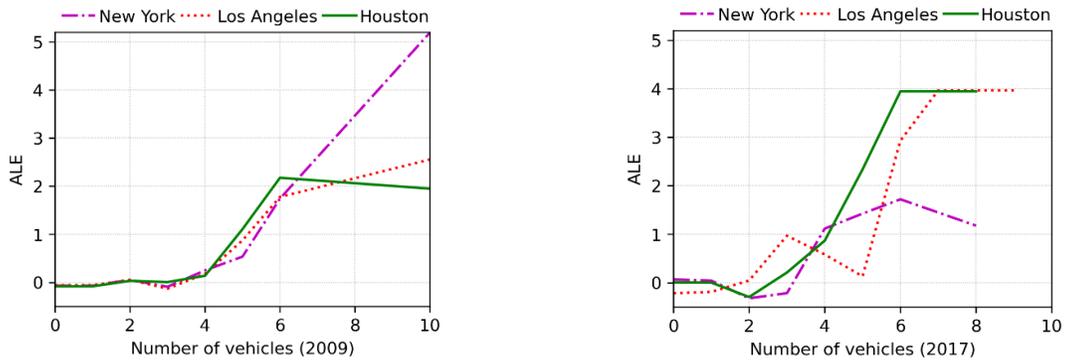

**Fig. 10.** ALE plot for the number of vehicles in a household of the three cities

Fig. 11 suggests that online purchases are almost not sensitive to the percentage of household members with a bachelor's degree in the three cities in 2009, which is quite different from the national trend. A possible explanation is that greater human-to-human interaction opportunities in big cities allow people even without higher education to learn and get familiar with shopping over the Internet, which largely annihilates the effect of higher education in online shopping. In 2017, the overall trends for the number of online purchases become more diverse beyond about 25% of household members having a bachelor's degree: in New York household online purchases decreases with a higher percentage of household members receiving college education, whereas the trends are largely the opposite in Los Angeles and Houston.

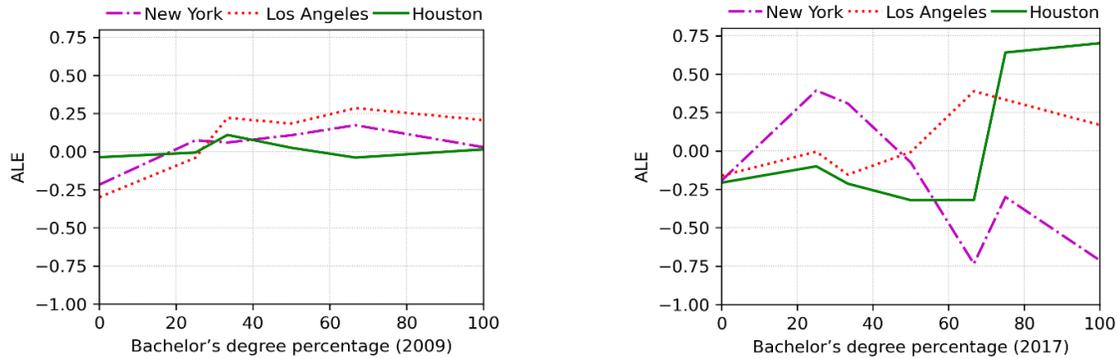

**Fig. 11.** ALE plot for bachelor's degree percentage in a household of the three cities

It is very interesting to see the opposite effect of home ownership in big cities compared to the national-level effect, as shown in Fig. 12 (compared to Fig. 4). In all three cities and in both years, with the exception of Houston in 2017, households with a renting home actually purchase more online. A possible explanation is that these cities constantly welcome new comers (college graduates and people moving in for jobs) who start their lives in these cities by first renting a place and need to purchase lots of items. Also, in these big cities rented and owned property types may be more similar (e.g., mostly apartment units/condos) than at the national level (renting an apartment unit vs. owning a single-family house). Thus, the potential property size effect associated with home ownership is diminished. The absolute value of the difference between renting vs. owning is also much larger for the big cities than for the national average, with the largest difference being 0.12 in Los Angeles (vs. about 0.007 nationally) in 2009, and nearly 0.14 in New York (vs. about 0.04 nationally) in 2017, which also corroborates the more active online shopping of households in these big cities.

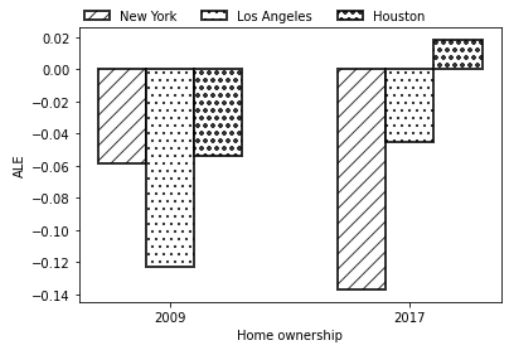

**Fig. 12.** ALE plot for home ownership of the three cities

The impacts of travel time that household members spend in the three cities are also somewhat different from the national level impacts. On the left-hand side of Fig. 13, while the decline of online purchases when travel time increases from zero is in line with that in Fig. 5, the increase of online purchases with travel time stops at around 275 min. What is also different from the national plot is that the overall variation in online purchases is less significant in 2017 than in 2009 in the three cities. In 2017, online purchases stay rather constant in the three cities when travel time per day goes over 300 min.

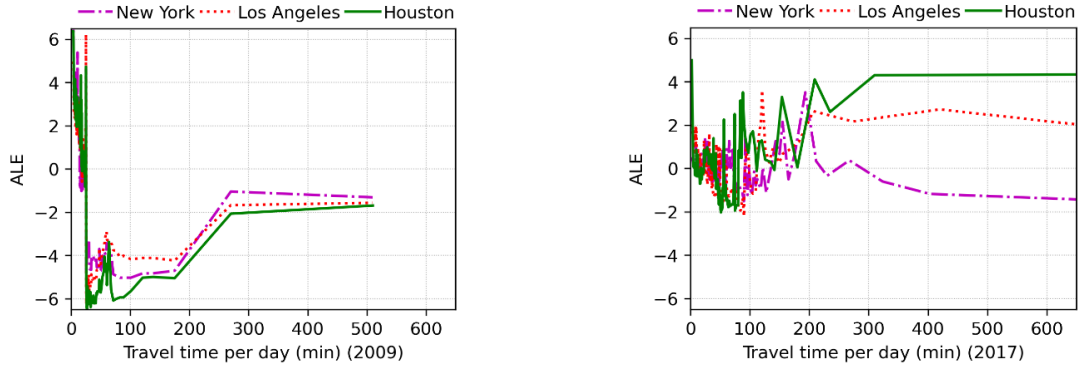

**Fig. 13.** ALE plot for travel time of household members per day of the three cities

For land use related variables, Fig. 14 shows that unlike the national-level trend, online purchases are not much sensitive to population density when the density of the area is high (above 17,000 people/sq. mile in both years). Fig. 15 further illustrates that the difference of living in urban and non-urban areas becomes much less in 2017 than in 2009, which again differs from the national trend and is a sign that online shopping becomes more prevalent across different areas in each of the cities. A possible explanation is that the proximity and relatively small socioeconomic and geographical differences of urban and "non-urban" (i.e., suburban) areas in each city (at the national level, "non-urban" can include really remote rural areas that have very different characteristics from urban areas) facilitate the spread of online shopping within each city. It is interesting to see that for Los Angeles, households in non-urban areas shop even more online than in urban areas.

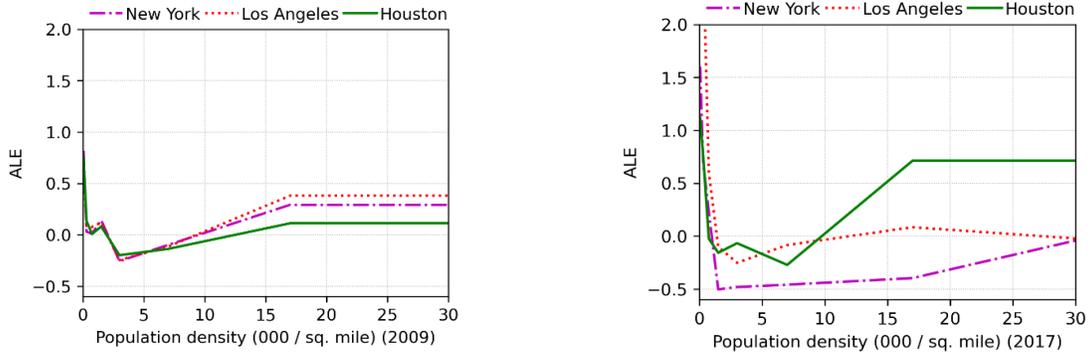

**Fig. 14.** ALE plot for population density of the three cities

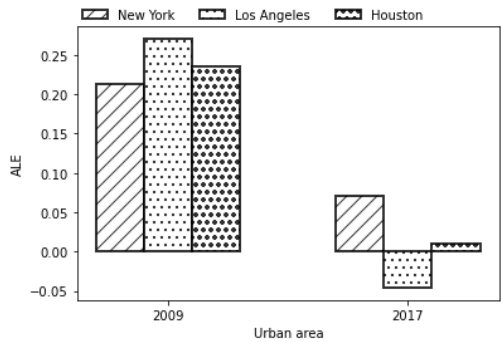

**Fig. 15.** ALE plot for urban area of the three cities

# 8 Conclusions

While online shopping behavior has been quite extensively studied in the existing literature, national-level investigation with a focus on predictive modeling and analysis remains limited. Different from the existing studies, this paper leverages the two most recent releases of the NHTS data in the U.S. to develop ML models, specifically GBM to predict online shopping purchases with extensive comparative analysis of the modeling results between 2009 and 2017. The NHTS data allow us not only to conduct national-level investigation but also at the household level, which is more appropriate than at the individual level given the connected consumption and shopping needs of members in a household. The comparative analysis includes quantifying the importance of each input variable in predicting online shopping demand, and characterizing the relationships between the predicted online shopping demand and the input variables, with the relationships flexible enough that can vary with the values of the input variables. The modeling employs a systematic procedure based on Recursive Feature Elimination algorithm to reduce the risk of model overfitting and increase model explainability. In performing the analysis, two latest advances in ML techniques, Shapley value-based feature importance and Accumulated Local Effects plots, are adopted which overcome the drawback of the prevalent techniques.

The modeling results show that GBM yields much higher prediction accuracy than several other ML (including regression) models. We find that household income contributes the most to predicting online shopping demand. Over time, the importance of Internet use and gender diminishes, while household member age and household size become more important. By employing the ALE technique, value-dependent effects of the input variables on predicted online shopping demand are estimated, which provide richer insights than single-number estimates as in prior research. The estimates show that the effect of the percentage of household members receiving higher education is not monotonic. The generation that grew up with online shopping significantly influence the effect of adult percentage in a household. Households owning home property tend to buy more online than if renting a living place. Total travel time of a household has an overall positive relationship with online purchases. However, the number of trips has a non-monotonic effect, with an explanation that more trips not only reduce the available time for shopping but also increase the chance of buying things on the way. The ALE plot for shopping trip percentage provides a mixed effect, suggesting that complementary and substitution relationships may both exist between online and in-store shopping. The relationship between population density of the living neighborhood and online purchases follows a "V" shape with plausible influencing factors being in-store shopping distance, social interactions, and Internet access. Living in an urban area and having daily Internet use encourage online shopping. As online shopping becomes more prevalent over time, the ALE plots further reveal the differences between 2009 and 2017. We also look into online shopping demand in three of the largest cities in the U.S., and discover commonalities with the national-level results as well as some unique characteristics for these cities.

This paper presents a beginning of taking a machine learning approach for predicting household-level online shopping demand, and for revealing the importance of influencing factors and their relationships with the demand. The models developed and insights gained can be used for online shopping-related freight demand generation and may also be considered for evaluating the potential impact on online shopping demand of relevant policies, e.g., land use planning, gasoline pricing, and transportation demand management to reduce trip-making. The proposed modeling approach could be further used as future releases of NTHS or similar data become available, which will help gain more in-depth understanding of the evolution of input variable importance and their relationships with household online shopping demand. The modeling and analysis could be extended with more advanced approaches, e.g., by combining GBM and a support vector classifier which first classifies household locations so that even higher prediction accuracy could be achieved.

# Acknowledgement

This research is supported in part by the U.S. National Science Foundation and the U.S. Department of Energy through the Argonne National Laboratory. Opinions expressed herein do not necessarily reflect those of the two agencies.

# Appendix 1: Candidate input variables for 2009

**Socioeconomic characteristics**

1. Average age of the household
2. Education
   - Percentage of members in the household not having high school degree
   - Percentage of members in the household having only high school
   - Percentage of members in the household having bachelor's degree
   - Percentage of members in the household having graduate degree
3. Percentage of male in the household
4. Percentage of race in the household
   - White
   - Black or African American
   - Asian
   - American Indian or Alaska Native
   - Native Hawaiian or other Pacific Islander
   - Multiple race
   - Some other race
5. Number of vehicles in the household
6. Household income
7. Household size
8. Percentage of workers in the household
9. Percentage of drivers in the household
10. Percentage of full-time worker in the household
11. Percentage of part time worker in the household
12. Percentage of adults in the household

**Trip characteristics**

1. Total travel time of the household member
2. Gas price
3. Number of total trips of the household
4. Percentage of shopping trips

**Land use characteristics**

1. Population density
2. Household area (categorical variables)

- Urban
- Suburban
- Rural

3. 2010 Census division classification for the household's home address (categorical variables)
   - New England
   - Middle Atlantic
   - East North Central
   - West North Central
   - South Atlantic
   - East South Central
   - West South Central
   - Mountain
   - Pacific

4. Home Ownership

5. House type
   - Duplex
   - Townhouse
   - Apartment or condominium
   - Mobile home or trailer

**Internet use**

1. Frequency of using the Internet (categorical variables)
   - All household members use the Internet daily
   - All household members use the Internet a few times a week
   - All household members use the Internet once in a week
   - All household members use the Internet a few times a month
   - No household member ever uses the Internet

# Appendix 2: Candidate input variables for 2017

**Socioeconomic characteristics**

1. Average age of the household
2. Education
   - Percentage of members in the household not having high school degree
   - Percentage of members in the household having only high school
   - Percentage of members in the household having bachelor's degree
   - Percentage of members in the household having graduate degree
3. Percentage of male in the household
4. Health status of the individual
   - Percentage of members in the household having excellent health condition
   - Percentage of members in the household having poor health condition
5. Percentage of race in the household
   - White
   - Black or African American
   - Asian
   - American Indian or Alaska Native
   - Native Hawaiian or other Pacific Islander
   - Multiple race
   - Some other race
6. Number of vehicles in the household
7. Household income
8. Household size
9. Percentage of workers in the household
10. Percentage of drivers in the household
11. Percentage of full-time worker in the household
12. Percentage of part time worker in the household
13. Percentage of adults in the household

**Trip characteristics**

1. Total travel time of the household member
2. Gas price
3. Number of total trips of the household
4. Percentage of shopping trips

**Land use characteristics**

1. Population density

2. Household area (categorical variables)

   o Urban

   o Suburban

   o Rural

3. 2010 Census division classification for the household's home address (categorical variables)

   o New England

   o Middle Atlantic

   o East North Central

   o West North Central

   o South Atlantic

   o East South Central

   o West South Central

   o Mountain

   o Pacific

4. Home Ownership

**Internet use**

1. Frequency of using the Internet (categorical variables)

   o All household members use the Internet daily

   o All household members use the Internet a few times a week

   o All household members use the Internet once in a week

   o All household members use the Internet a few times a month

   o No household member ever uses the Internet

2. Household that use smartphone daily

3. Household that use laptop/desktop daily

# Appendix 3: Correlation matrices of the input variables

## 2009

| | Average member age | Male percentage | Household size | Household income | Adult percentage | No high school percentage | Bachelor's degree percentage | Number of vehicles | Home ownership | Number of trips per day | Travel time per day | Gas price | Shopping trip percentage | Urban area | Population density | Daily Internet use |
|---|---|---|---|---|---|---|---|---|---|---|---|---|---|---|---|---|
| Average member age | 1.00 | | | | | | | | | | | | | | | |
| Male percentage | -0.01 | 1.00 | | | | | | | | | | | | | | |
| Household size | -0.54 | 0.06 | 1.00 | | | | | | | | | | | | | |
| Household income | -0.13 | 0.13 | 0.19 | 1.00 | | | | | | | | | | | | |
| Adult percentage | 0.51 | 0.02 | -0.77 | -0.13 | 1.00 | | | | | | | | | | | |
| No high school percentage | -0.06 | 0.01 | 0.07 | -0.17 | -0.03 | 1.00 | | | | | | | | | | |
| Bachelor's degree percentage | -0.05 | 0.04 | 0.02 | 0.20 | -0.07 | -0.11 | 1.00 | | | | | | | | | |
| Number of vehicles | -0.20 | 0.10 | 0.34 | 0.30 | -0.08 | -0.02 | -0.01 | 1.00 | | | | | | | | |
| Home ownership | 0.15 | 0.01 | 0.02 | 0.25 | 0.06 | -0.09 | 0.05 | 0.22 | 1.00 | | | | | | | |
| Number of trips per day | -0.16 | 0.07 | 0.28 | 0.24 | -0.15 | -0.03 | 0.07 | 0.22 | 0.07 | 1.00 | | | | | | |
| Travel time per day | 0.00 | 0.03 | -0.01 | 0.03 | 0.03 | 0.00 | -0.01 | 0.04 | 0.00 | -0.14 | 1.00 | | | | | |
| Gas price | -0.01 | 0.00 | 0.00 | -0.01 | 0.01 | 0.00 | -0.02 | 0.00 | -0.01 | 0.01 | 0.01 | 1.00 | | | | |
| Shopping trip percentage | 0.17 | 0.03 | -0.13 | -0.10 | 0.15 | 0.01 | -0.02 | -0.06 | -0.01 | -0.07 | -0.13 | 0.00 | 1.00 | | | |
| Urban area | -0.03 | 0.01 | -0.02 | -0.03 | 0.01 | 0.01 | 0.01 | -0.13 | -0.14 | -0.01 | -0.01 | 0.01 | 0.01 | 1.00 | | |
| Population density | -0.05 | 0.01 | -0.02 | -0.02 | 0.01 | 0.01 | 0.02 | -0.19 | -0.21 | -0.01 | -0.01 | 0.01 | 0.01 | 0.63 | 1.00 | |
| Daily Internet use | -0.11 | 0.07 | 0.10 | 0.28 | -0.08 | -0.11 | 0.11 | 0.08 | 0.06 | 0.17 | 0.00 | -0.02 | -0.04 | 0.00 | 0.01 | 1.00 |

| | Average member age | Male percentage | Household size | Household income | Adult percentage | No high school percentage | Bachelor's degree percentage | Number of vehicles | Home ownership | Number of trips per day | Travel time per day | Gas price | Shopping trip percentage | Urban area | Population density | Daily Internet use |
|---|---|---|---|---|---|---|---|---|---|---|---|---|---|---|---|---|
| Average member age | 1.00 | | | | | | | | | | | | | | | |
| Male percentage | -0.08 | 1.00 | | | | | | | | | | | | | | |
| Household size | -0.22 | 0.11 | 1.00 | | | | | | | | | | | | | |
| Household income | -0.13 | 0.12 | 0.34 | 1.00 | | | | | | | | | | | | |
| Adult percentage | 0.20 | 0.00 | -0.28 | -0.06 | 1.00 | | | | | | | | | | | |
| No high school percentage | 0.00 | 0.00 | 0.08 | -0.14 | -0.30 | 1.00 | | | | | | | | | | |
| Bachelor's degree percentage | -0.15 | 0.03 | -0.01 | 0.15 | 0.03 | -0.12 | 1.00 | | | | | | | | | |
| Number of vehicles | -0.10 | 0.17 | 0.55 | 0.36 | -0.10 | -0.04 | 0.01 | 1.00 | | | | | | | | |
| Home ownership | 0.29 | 0.01 | 0.21 | 0.27 | -0.01 | -0.07 | -0.01 | 0.33 | 1.00 | | | | | | | |
| Number of trips per day | -0.12 | 0.05 | 0.60 | 0.25 | -0.17 | 0.02 | 0.03 | 0.35 | 0.15 | 1.00 | | | | | | |
| Travel time per day | -0.09 | 0.06 | 0.33 | 0.15 | -0.08 | 0.03 | 0.00 | 0.21 | 0.07 | 0.04 | 1.00 | | | | | |
| Gas price | 0.04 | 0.00 | 0.00 | 0.05 | 0.01 | -0.03 | -0.01 | -0.01 | -0.04 | 0.00 | 0.01 | 1.00 | | | | |
| Shopping trip percentage | 0.16 | 0.03 | -0.06 | -0.10 | 0.04 | 0.04 | -0.05 | -0.07 | 0.00 | -0.20 | -0.06 | 0.00 | 1.00 | | | |
| Urban area | -0.12 | 0.00 | -0.06 | 0.04 | 0.02 | 0.00 | 0.05 | -0.15 | -0.18 | -0.04 | -0.01 | 0.18 | -0.01 | 1.00 | | |
| Population density | -0.17 | 0.00 | -0.08 | 0.05 | 0.02 | -0.01 | 0.08 | -0.22 | -0.25 | -0.05 | -0.02 | 0.17 | -0.01 | 0.64 | 1.00 | |
| Daily Internet use | -0.28 | 0.02 | 0.15 | 0.27 | -0.04 | -0.17 | 0.13 | 0.14 | 0.03 | 0.12 | 0.06 | 0.02 | -0.10 | 0.03 | 0.06 | 1.00 |